\documentclass[letterpaper, 10 pt, conference]{ieeeconf}  
\usepackage{graphicx}
\usepackage{amsmath}
\usepackage{amssymb}
\usepackage{verbatim} 
\usepackage{xcolor}
\DeclareUnicodeCharacter{0308}{HERE!HERE!}

\IEEEoverridecommandlockouts                              

\title{\LARGE \bf Height Estimation of Children under Five Years using Depth Images}
\author{Anusua Trivedi$^{1}$,  Mohit Jain$^{1}$, Nikhil Kumar Gupta$^{2}$, Markus Hinsche$^{2}$, Prashant Singh$^{2}$\\
Markus Matiaschek$^{2}$, Tristan Behrens$^{2}$,
Mirco Militeri$^{1}$, Cameron Birge$^{1}$, Shivangi Kaushik$^{2}$\\
Archisman Mohapatra$^{3}$, Rita Chatterjee$^{4}$, Rahul Dodhia$^{1}$, Juan Lavista Ferres$^{1}$
\thanks{$^{1}$Microsoft {\tt\small\{antriv, mohja\}@microsoft.com}}
\thanks{$^{2}$Child Growth Monitor, Welthungerhilfe}
\thanks{$^{3}$Executive Director at GRID Council, India}
\thanks{$^{4}$(Retired) Professor of Pediatrics, Dr. B C Roy PGI PS, India}
}
\begin{document}
\maketitle
\thispagestyle{empty}
\pagestyle{empty}
\begin{abstract}

Malnutrition is a global health crisis and is a leading cause of death among children under 5 years. Detecting malnutrition requires anthropometric measurements of weight, height, and middle-upper arm circumference. However, measuring them accurately is a challenge, especially in the global south, due to limited resources. In this work, we propose a CNN-based approach to estimate the height of standing children under 5 years from depth images collected using a smartphone. According to the SMART Methodology Manual, the acceptable accuracy for height is less than 1.4 cm. On training our deep learning model on 87131 depth images, our model achieved a mean absolute error of 1.64\% on 57064 test images. For 70.3\% test images, we estimated height accurately within the acceptable 1.4 cm range. Thus, our proposed solution can accurately detect stunting (low height-for-age) in standing children below 5 years of age.
\end{abstract}
\section{Introduction}
Malnutrition refers to an imbalance of nutrition, both under and over-nutrition. According to the World Health Organization (WHO), malnutrition is a global health crisis and is the reason behind $\sim$45\% of deaths among children under 5~\cite{who-malnutrition}. The risk of malnutrition is especially high among children below the age of 5, and effective early interventions can help overcome the alarming situation. Malnutrition is categorized into undernutrition, micro-nutrient-related malnutrition, and overweight. Deficiency of nutrition, i.e. \textit{undernutrition}, is the leading reason for malnutrition in the global south. It is mainly associated with poor socioeconomic conditions due to unavailability of enough food to eat, infectious diseases, and/or lack of knowledge about young child care. Undernutrition makes the children more vulnerable to other diseases as well, and increases the risk of death. There are three forms of undernutrition: wasting (low weight-for-height), stunting (low height-for-age) and underweight (low weight-for-age).

Measuring malnutrition accurately is challenging.
Early detection and intervention require regular anthropometric measurements, measuring weight, height, and middle-upper arm circumference of children under 5. However, such measurements may also have errors due to inexpert data collectors, inadequate tools, and/or poor data management~\cite{commcare}. According to SMART Methodology Manual~\cite{smart}, the acceptable technical error of measurement (\textit{i.e.}, variance between two rounds of height measurement) is less than 1.2 cm, and the acceptable accuracy (\textit{i.e.}, difference between the measured height and ground truth) is less than 1.4 cm. 

In this paper, we propose a Convolutional Neural Network (CNN) based method to accurately estimate the height of standing children under 5 from depth images collected using a commercial off-the-shelf smartphone. Overall, we collected data of 3887 children (2581 train data, 1306 test data) aged 2-5 years in rural India. Our approach estimated height with a mean absolute error of 1.64\%, and for 70.3\% test images, it achieved the acceptable 1.4 cm range. Hence, our solution can detect stunting accurately, by predicting the estimated height with the child's age.
\begin{figure*}
 \centering
  \includegraphics[height=4cm]{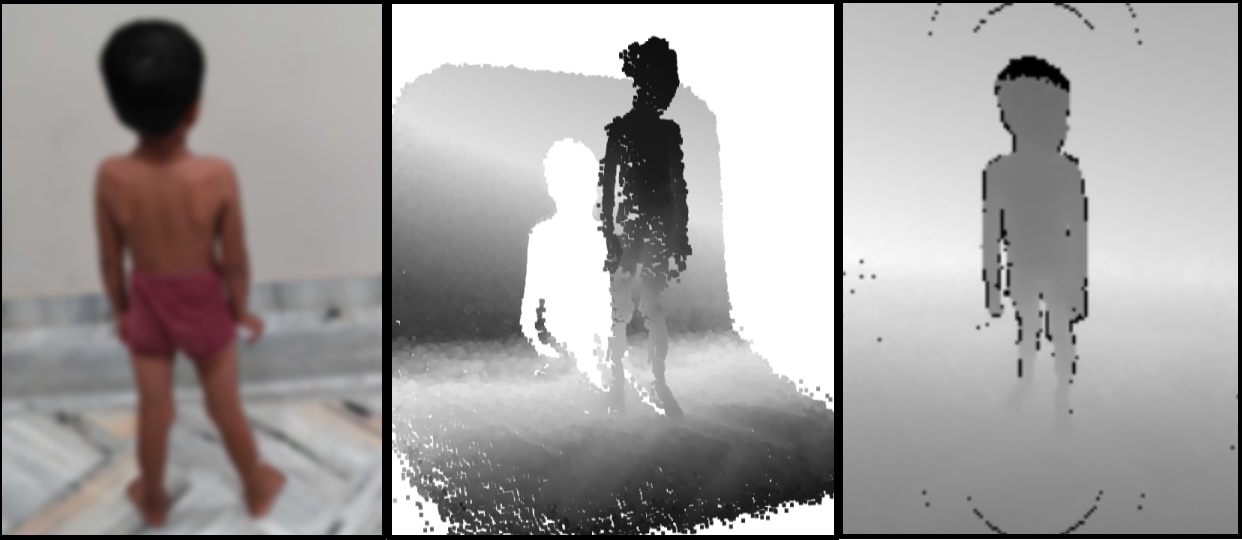}
  \caption{(from left to right) (a) Back video of a child. (b) Point cloud data. (c) Depth image.}\label{fig:data_types}
\end{figure*}

\section{Related Work}
Estimating anthropometric measurements, especially height, using images has been an active area of research~\cite{height-singleview, height-plosone, height-cnn}. Researchers have proposed height estimation using a single image~\cite{height-singleview}, images of a subject taken from multiple views~\cite{height-multiviewimages, height-plosone}, and from three-dimensional depth images~\cite{height-cnn}. A single image of the subject requires a cubical reference object of known dimension along with complex calibration and genetic algorithm~\cite{height-singleview}. The proposed approach was only evaluated with one subject and achieved a high estimation error of 5.5\%. Multiple views (usually 5 or more views) of the subject enables three-dimensional reconstruction of the body surface to estimate height and other anthropometric measurements. Li \textit{et al}.~\cite{height-multiviewimages} evaluated their multi-view approach with a child mannequin, while Liu \textit{et al}.~\cite{height-plosone} validated with 31 adults to achieve $\sim$1\% error. More recently, with easy access to depth cameras in Microsoft Kinect and Google Tango devices, depth images have been used for anthropometric measurements. Yin and Zhou~\cite{height-cnn} used a single depth image and passed it through a four-stage CNN to predict lengths of different body parts and total height. The proposed method was executed on 2136 images collected from 14 adults in 10 different postures. Overall, their method performed well across postures, achieving a total average error of 0.9\%, and they found depth images to outperform RGB images. However, in all the above cases, these papers proposed methods to measure height of adults only. To the best of our knowledge, none of the prior approaches were tested with standing children below 5 years, which is the key novelty of our work.

\section{Dataset}
All the data was collected from two states of India, Rajasthan (Baran district) and Madhya Pradesh (Chatarpur and Sheopur districts), during 2017-2019, using the Child Growth Monitor~\cite{CGM} phone app developed by Welthungerhilfe.
We focus on data collected for children who can stand (usually 2-5 years of age) for this work. The data was collected in the regional Anganwadi centers. Anganwadi is a type of rural child care center in India.
The data collectors were mostly young adults (20-30 years old) and received a four-day data collection training.

After getting consent from their parents/grandparents, children were asked to stand in front of a solid-colored wall. If needed, a white banner was placed behind the child to replicate a wall. All the videos were recorded using the Lenovo Phab 2 Pro phone, which has a time-of-flight sensor to capture \textit{point cloud data} at 1920x1080 resolution with three frames/second.
The point cloud videos were converted into depth images in the data processing stage. For each child, the data collector used the phone app to collect three point cloud videos: (a) \textit{front video}: where the child is facing the camera, (b) \textit{back video}: where the child's back is facing the camera (Figure~\ref{fig:data_types}a), and (c) 360-degree video: where the child was asked to spin slowly to capture a 360-degree view of the child. The data collector decided the length of these videos; usually, the front and back videos were 2-4 seconds long, while the 360-degree videos were 5-8 seconds. (Note: For front and back data, a single image would have sufficed, however as children move frequently, we opted for videos.) The data collector ensured that the child's head to toe was fully visible in each video. Next, manual measurements of the ground truth weight, height, and mid-upper arm circumference (MUAC) were taken, using the standardized weight machine, height board, and MUAC tape, respectively. On average, it took 15-20 mins to collect data for a child, involving consent forms, digital videos, and manual measurements. In case the child did not co-operate, they moved to the next child. The child and/or guardian did not receive any incentive for participation. 

Overall, data was collected for 3887 children, and the age-wise distribution of the point cloud video dataset is shown in Table~\ref{tab:tab1}.

Note: The data collection was performed in compliance with the Indian Council of Medical Research 2017 guidelines for biomedical research involving human participants, ensuring the core principles of ethics -- autonomy, beneficence, non-maleficence and justice.
Participant anonymity, confidentiality and privacy was maintained at all stages.

\section{Data Transformations}
\begin{figure}
  \includegraphics[width=\linewidth]{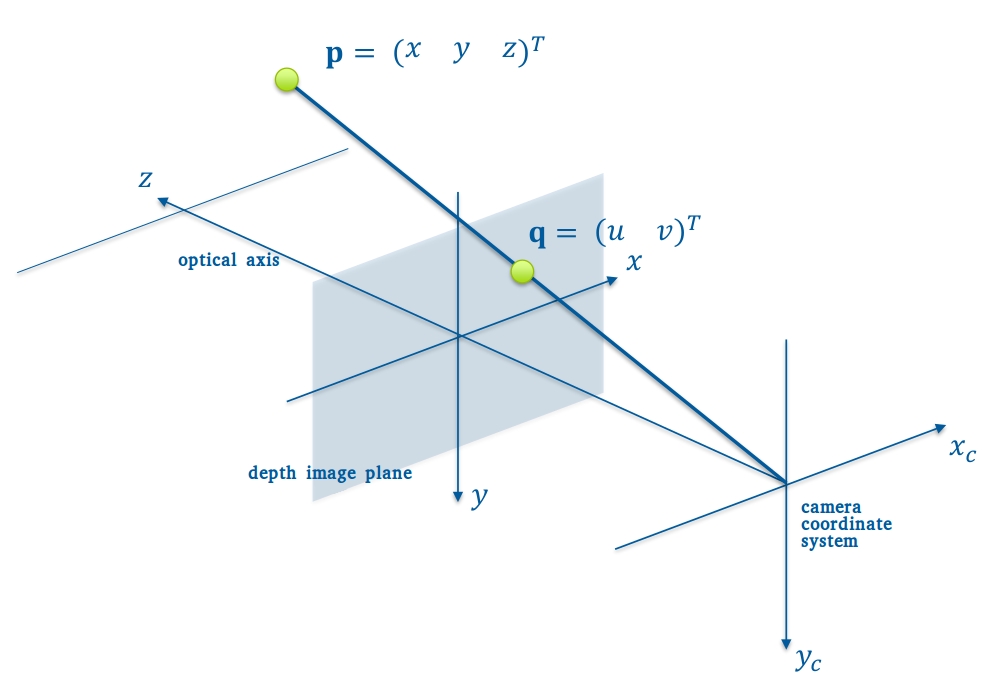}
  \caption{Point cloud to Depth Image Transformation}\label{fig:pcd_to_depthmap}
\end{figure}

Point cloud data is a flexible file format often used to store multidimensional data. In our case, the point cloud data constitutes of a set of points, wherein a single point $p$ represents the three components of three-dimensional space, \textit{i.e.}, $x$, $y$, and $z$ values (visualized in Figure~\ref{fig:data_types}b and  Figure~\ref{fig:pcd_to_depthmap}). Each collected point cloud video comprises several image frames. As part of data transformation, we extract each point cloud image frame and convert them to depth images, thus obtaining our dataset of 144195 depth images (Table~\ref{tab:tab2}). Note: As the child is always moving, each image frame is (slightly) different, thus helping us to augment our dataset naturally.
The video type---front, back, and 360-degree---based distribution of our depth images is shown in Table~\ref{tab:tab2}.

A \textit{depth image} is an image that contains information related to the distance of the surfaces of objects from a viewpoint in the real world.
For each pixel in a depth image, it has a $d$ distance value (`depth') from the camera sensor coordinate system to the object. Hence, a depth image is represented as a matrix, where each cell (or pixel) contains a single metric depth information (shown in Figure~\ref{fig:data_types}c), accessed by two coordinates, $u$ and $v$ (Figure~\ref{fig:pcd_to_depthmap}).

This transformation from point cloud data to depth image is performed using a projection matrix. The parameters in the projection matrix depend upon the camera specification, including sensor size and focal length. For every point $p$ in the point cloud, there exists a point $q$ in the depth image plane (Figure~\ref{fig:pcd_to_depthmap}). In other words, the pixel on the depth image plane is a projection of the real world. (Note: Latest phones with depth cameras provide direct access to depth images; hence this data transformation step is not needed.)

As the data was collected for children below the age of 5, it was noisy with several frames in a point cloud video consisting of poor quality data. Hence, the videos were graded manually based on their quality. A video with any frame consisting of noisy data, \textit{i.e.}, no child visible, blurry, too dark, several people present, \textit{etc.}, was rated `\textit{bad}'. Only `\textit{good}' videos were included in the test dataset, resulting in videos from 1306 children, with 57064 depth images (Table~\ref{tab:tab2}).

\begin{table}
\caption{Age-wise distribution of training and test videos} 
\centering 
\begin{tabular}{c|c|rr} 
\hline\hline 
\textbf{Age} & \textbf{Total} & \textbf{Training} & \textbf{Test}\\ [0.25ex] 
\hline   
2-3 & 1030 & 712 & 318\\
3-4 & 1370 & 895 & 475\\
4-5 & 1487 & 974 & 513\\
\hline
\textbf{Total} & \textbf{3887} & 2581 & 1306\\
\hline\hline  
\end{tabular} 
\label{tab:tab1} 
\end{table}

\begin{table}
\caption{Video type distribution of obtained depth images}
\centering 
\begin{tabular}{c|c|rr} 
\hline\hline 
\textbf{Video Type} & \textbf{Total} & \textbf{Training} & \textbf{Test}\\ [0.25ex] 
\hline
\hline
Front video & 38602 & 24852 & 13750\\
Back video & 37333 & 23272 & 14061\\
360-degree video & 68260 & 39007 & 29253\\
\hline
\textbf{Total} & \textbf{144195} & 87131 & 57064\\
\hline\hline
\end{tabular}
\label{tab:tab2}
\end{table}

\section{Method}
Our data collector collected three videos of each child using the smartphone, along with manual anthropometric measurements, including the height of the children. We used the point cloud data and manual height labels to train various deep learning models, such as GAPNet and PointNet, focusing on minimizing the mean absolute error in height estimation.

PointNet~\cite{8099499} uses raw point cloud data as input without any emphasis on their ordering, while GAPNet~\cite{DBLP:journals/corr/abs-1905-08705} exploits local features by introducing GAPLayer, which assigns different attention weights on the neighborhood for each point. We trained these models on our point cloud data, however even after tuning, the mean absolute error was very high at 4 cm, which was unacceptable. This led us to transform the point cloud data to depth images, which was used to train a Convolutional Neural Network (CNN) based deep learning model for height estimation. Our model consists of 12 convolutional and three dense layers, with padding of 240x180 pixels. We used Rectified Linear Unit (ReLU) as activation function and Mean Squared Error as the loss function.

\section{Evaluation Metrics}
Apart from the mean absolute error and mean absolute percentage error on the test dataset, we also evaluated our approach using the Standardisation Test evaluation metric~\cite{smart}.
A standardisation test for anthropometric measurements is a practical assessment of the data collector's measurement skills. It helps to objectively evaluate the quality---precision, and accuracy---of the measurements taken by each data collector. The standardisation test consists of each collector measuring ten different children twice in two rounds of measurements. Along with the standardisation test for manual measurements, we adapted the standardisation test for our point cloud data. For that, each data collector captured three videos of the child at each round of measurement as well.

For evaluation, we calculated two metrics:
\begin{itemize}
\item\textbf{Intra TEM (Technical Error of Measurement)} is used to evaluate the measure of \textit{precision} in the standardisation test. For each data collector, it is calculated using the variance between the height predictions on the videos of a child taken in two rounds of measurement.
\item\textbf{Bias from Supervisor} is used to evaluate the measure of \textit{accuracy} in the standardisation test. For each data collector, it is calculated using the mean of height predictions using the videos of a child compared to the ground truth (\textit{i.e.}, an expert supervisor's manual measurements).
\end{itemize}

We use guiding principles from the SMART (Standardized Monitoring and Assessment of Relief and Transitions) Methodology Manual ~\cite{smart} to compare the manual versus our models' measurements.
Table~\ref{table:smart-range} shows acceptable limits for Intra TEM and Bias from Supervisor in a standardisation test. According to this table, any Intra TEM prediction with less than 1.2 cm and any Bias from Supervisor prediction with less than 1.4 cm are acceptable.

\begin{table}
\centering
\caption{SMART Manual Measurement ranges}
\begin{tabular}{c | c c} 
 \hline\hline
 & \textbf{Intra TEM Range} & \textbf{Bias from Supervisor Range}\\ [0.5ex] 
 \hline
 Good & \(<0.4 cm\) & \(<0.4 cm\)\\ 
 Fair & \(<0.6 cm\) & \(<0.6 cm\)\\
 Poor & \(<1.2 cm\) & \(<1.4 cm\)\\
 Reject & \(>1.2 cm\) & \(>1.4 cm\)\\
 \hline\hline
\end{tabular}
\label{table:smart-range}
\end{table}

\begin{table}
\centering
\caption{Performance of our CNN model on test dataset}
\begin{tabular}{ c | l l} 
\hline\hline
 \textbf{Video Type} & \textbf{In 1.4 cm Range} & \textbf{MAPE} \\ [0.5ex] 
 \hline
 Front & 71.77\% & 1.586\% \\  
 Back  & 69.48\% & 1.671\% \\
 360-degree   & 69.61\% & 1.680\% \\
 \hline
 \textbf{Average}  & \textbf{70.28\%} & \textbf{1.645\%} \\ 
 \hline\hline
\end{tabular}
\label{table:cnneval}
\end{table}

\section{Evaluation of our Model}
Our CNN model achieved a mean absolute error of 1.4 cm and a MAPE (mean absolute percentage error) of 1.64\%. Table~\ref{table:cnneval} shows the results on the test dataset using our CNN model, with 71.77\% of front videos, 69.48\% of back videos, and 69.61\% of 360-degree videos were under 1.4 cm acceptable range of height estimation, as per SMART Methodology Manual~\cite{smart}. We see that the font videos perform the best, even though 360-degree videos offer more surface area. This result is in alignment to the work of Hung et. al. ~\ref{hung2004anthropometric}, which shows the front scans are more effective than circumference scans for anthropometric measures. As the test and training set were completely disjoint, it shows that our proposed CNN model works reasonably well on unseen real-world data. Moreover, we collected ten children data (5 male and 5 female, with an average age of 5$\pm$1.5 years) using six data collectors and an expert supervisor's ground truth height labels. We use our CNN model to predict the height of the children using the captured front videos (as among the three video types, the front videos performed slightly better) by the data collectors. This standardisation test data was collected during Jan'2020 in Rajasthan, India.

With respect to Intra TEM, comparing manual measurements, all six data collectors performed in the `Good' range on an average. In contrast, for our CNN model, four data collectors were in the `Poor' range, and two were in the `Reject' range.

With respect to Bias for Supervisor, comparing manual measurements, all the data collectors performed in the `Good' range, and for our CNN model, all data collectors were in the `Poor' range, thus were acceptable too.  This shows that our CNN model achieves accuracy but lacks precision.

\section{Discussion}
One or more forms of malnutrition affect every country in the world. Combating malnutrition in all its forms is one of the most significant challenges for global health. One of the big problems in tackling malnutrition is that it is difficult to identify by conventional means or naked eye whether a child is suffering from malnutrition. Due to flawed data, most of the time, the field-aid workers cannot reach out to children who urgently require assistance. Anthropometric measurements defining malnutrition may not be without errors if taken by inexpert hands and/or inadequate tools. Thus an enhanced ability to improve anthropometric measures in children and determine an individual's propensity to develop or have progressive complications from malnutrition would be of enormous benefit. Recent technological growth has allowed the development of clinical data acquisition and analysis of such data much more straightforward. Efficient and effective smartphone devices will enable us to non-invasively collect more informative images (like point cloud data and depth images) in large populations. In this paper, we show the benefits of combining advanced imaging technology with deep learning methods for estimating height (which can be used to predict stunting) in standing children below 5 years of age. The availability of complex images combined with the rapid evolution of computational data science offers promising opportunities for extracting new inferences and actionable insights that have the potential to improve health outcomes significantly. This more sophisticated data-enriched environment, in turn, has the potential to allow better clinical decision-making through support by automated means, encouraging moves towards intelligent assistance and diagnosis.

\section{Ethics Statement}
Welthungerhilfe collected all the patient data for this study. All data is de-identified and anonymized. Consent for the use of this data for research was obtained from the participants. Institute Ethics Committee of All India Institute of Medical Sciences, Patna India (AIIMS IEC) approved our study.

{
\small
\bibliographystyle{ieee}
\bibliography{egbib}
}
\end{document}